\documentclass{article}

\usepackage[preprint]{neurips_2026}

\usepackage[utf8]{inputenc} 
\usepackage[T1]{fontenc}    
\usepackage{hyperref}       
\usepackage{url}            
\usepackage{booktabs}       
\usepackage{amsfonts}       
\usepackage{nicefrac}       
\usepackage{microtype}      
\usepackage{xcolor}         
\usepackage{amsmath}
\usepackage{amssymb}
\usepackage{algorithm}
\usepackage{algpseudocode}
\usepackage{graphicx}
\usepackage{colortbl}
\usepackage{tikz}
\usepackage{pgfplots}
\pgfplotsset{compat=1.18}
\usepackage{authblk}
\usepackage{multirow}
\usepackage{wrapfig}
\usepackage{booktabs}
\usepackage[table]{xcolor}
\usetikzlibrary{shapes.geometric}
\usepackage{url}

\definecolor{best}{HTML}{CDE2CD} 
\definecolor{second}{HTML}{CFE2F3}

\title{TrioPose: Native Triple-Stream Diffusion Transformers for Pose-Guided Text-to-Image Generation}

\author{%
  Dian Gu \quad\quad Zhengyi Yang \\
  Institute of Automation, Chinese Academy of Sciences \\
  Beijing, China \\
  \texttt{\{gudian2024, zhengyi.yang\}@ia.ac.cn} 
}

\begin{document}

\maketitle

\begin{abstract}
Pose-guided text-to-image generation often suffers from limb distortions and feature crosstalk in complex multi-person scenarios. While existing UNet-based adapters struggle with long-range spatial dependencies, emerging Multimodal Diffusion Transformers (MM-DiTs) offer superior global modeling. However, naive signal concatenation in MM-DiTs severely disrupts pre-trained latent distributions. To address this, we propose TrioPose, a native pose-driven framework built upon the SD3.5M architecture. Specifically, we introduce a Triple-Stream Pose-Aware DiT (TSPA-DiT) that treats pose as an independent modality. It employs layer-wise activation and zero-initialized dual-residual injection to smoothly enforce geometric constraints while preserving pre-trained latent stability. To resolve severe multi-instance occlusions, we design a Learnable Relational Bias Mask that categorizes topological connectivity into fine-grained physical states, mapping them into continuous attention soft constraints to effectively decouple inter-instance interference. Furthermore, a Pose-Guided Spatial Loss Weighting strategy modulates the native diffusion objective using heatmap-derived error maps, focusing anatomical supervision strictly on distortion-prone regions. Extensive experiments demonstrate that TrioPose achieves state-of-the-art performance across challenging benchmarks, including Human-Art, CrowdPose, and OCHuman. Notably, it attains an AP of $64.33$ on Human-Art, representing a $30\%$ improvement over prior arts, while setting new standards for visual fidelity and text-image semantic alignment in complex multi-human generation.
\end{abstract}
\section{Introduction}
\label{sec:intro}

Text-to-image (T2I) diffusion models \cite{ho2020ddpm, rombach2022sd1.5} have transformed digital content generation. However, high-precision applications, such as animation production \cite{pan2024synthesizing}, virtual try-on \cite{yu2019vtnfp}, and embodied AI simulations, demand control beyond natural language, which often struggles to define the complex joint positions and fine-grained dynamics of deformable entities like humans \cite{ho2020ddpm}. Controllable human image generation (HIG) addresses this by aligning high-level semantics with structural priors, such as human pose \cite{ma2017pg2, xu2021text}, body parsing \cite{yu2019vtnfp, zhu2017your}, and multi-granularity language descriptions \cite{liu2022verbal, xu2021manipulation}. By lowering the technical barriers to high-quality content creation, these controllable synthesis techniques significantly streamline production workflows across creative industries.

While earlier HIG solutions based on Generative Adversarial Networks (GANs) \cite{goodfellow2014gan, lv2021learning, men2020controllable, zhang2021pise} and Variational Auto-encoders (VAEs) \cite{cheong2022pce, kingma2013vae} achieved success in specific domains, they often faced training instability and limited multi-modality fusion capabilities. The field has since transitioned to diffusion-based architectures \cite{ho2020ddpm, ramesh2022hierarchical}. Existing state-of-the-art methods \cite{ju2023humansd, zhang2023controlnet, zhao2023unicontrol, li2023gligen, mou2024t2i, wang2024stable, xuan2025spctrl, yin2025grpose} primarily rely on CNN-based UNet backbones, utilizing frozen weights and auxiliary adapters for spatial conditioning. Although effective for standard single-person synthesis \cite{xuan2025spctrl}, the local receptive field of convolutions limits global interaction modeling. Consequently, multi-person scenarios involving overlapping poses frequently lack long-range spatial dependencies, resulting in limb malformations and anatomical distortions.

Currently, the paradigm is shifting from UNet-based models to Multimodal Diffusion Transformers (MM-DiT) \cite{esser2024sd3}, which offer superior global context modeling. However, this transition challenges existing pose control strategies. Traditional plug-in concatenation \cite{mou2024t2i, wang2024stable} forcibly injects signals at shallow feature levels, which in MM-DiT architectures disrupts the aligned latent space, leading to convergence issues. Furthermore, in multi-subject interactions, global self-attention \cite{wang2024stable} induces signal crosstalk and identity confusion, while standard hard masking fails to capture true limb dependencies during complex overlaps.

To address these challenges, we propose TrioPose, a native pose modeling framework built upon the SD3.5M model \cite{esser2024sd3}. Discarding traditional adapters, we introduce the Triple-Stream Pose-Aware DiT (TSPA-DiT), treating pose as an independent modality parallel to text and image. To maintain latent space stability, TSPA-DiT employs Layer-wise Activation and Dual Residual Injection with zero-initialization, enabling a smooth transition from unconditional to constrained generation. To mitigate interference in multi-person interactions, we design a Learnable Relational Bias Mask that decouples topological connectivity into five physical states. These are mapped to continuous attention biases to adaptively regulate cross-regional interaction. Finally, we introduce Pose-Guided Spatial Loss Weighting, where a frozen ViTPose \cite{xu2023vitpose} extracts keypoint errors to re-weight the denoising objective, focusing anatomical supervision on distortion-prone regions. TrioPose achieves state-of-the-art performance, yielding a $64.33$ AP on Human-Art \cite{ju2023humanart}, outperforming prior art \cite{yin2025grpose} by $30\%$.
The main contributions are summarized as follows:
\begin{itemize}
\item We propose TSPA-DiT, a novel triple-stream architecture that encodes pose as an independent modality, utilizing zero-initialized Dual Residual Injection for stable, high-fidelity control without disrupting the MM-DiT distribution.
\item We design a Learnable Relational Bias Mask to map fine-grained physical states into continuous attention biases, effectively mitigating feature crosstalk in complex overlaps via soft constraints.
\item We introduce Pose-Guided Spatial Loss Weighting to transform anatomical supervision into latent spatial weights for the native denoising objective, thereby improving structural rigor.
\item Extensive experiments demonstrate that TrioPose achieves state-of-the-art performance, notably yielding an AP of $64.33$ on the HumanArt dataset, representing a $30\%$ improvement over existing methods.
\end{itemize}
\section{Related Works}
\label{sec:relate}

\paragraph{Controllable Diffusion Models:} Text-to-image diffusion models excel in visual synthesis but struggle with fine-grained spatial control using text alone. To address this, ControlNet \cite{zhang2023controlnet} introduces a dual-branch architecture with frozen pre-trained weights to encode spatial conditions for structure-aware generation. Similarly, T2I-Adapter \cite{mou2024t2i} and GLIGEN \cite{li2023gligen} offer efficient feature injection and layout constraints via lightweight adapters and gated self-attention, respectively. For complex multi-condition guidance, Uni-ControlNet \cite{zhao2023unicontrol} designs a unified multi-resolution adapter, while PixelPonder \cite{pan2025pixelponder} dynamically allocates computational priority to patches during inference, mitigating feature conflicts and accelerating generation. However, these adapter-based methods rely heavily on UNet architectures. Directly transferring these plug-and-play concatenation strategies to emerging MM-DiT severely disrupts the aligned latent distribution, causing convergence failures and structural distortions.

\paragraph{Pose-Guided Human Image Generation:} Pose-guided generation synthesizes human visuals based on structural priors. Early works like PG2 \cite{ma2017pg2} formulated this task as conditional image translation, though they suffered from pixel misalignment during large pose transfers. To handle spatial deformation, PATN \cite{zhu2019patn} leverages progressive attention, while Liquid Warping GAN \cite{goodfellow2014gan} and DensePose-based methods \cite{grigorev2019coordinate} utilize 3D flows and UV mapping to mitigate occlusion and texture loss. PISE \cite{zhang2021pise} further employs parsing maps for part-level decoupling and local texture injection. Diffusion models have advanced synthesis fidelity; specifically, HumanSD \cite{ju2023humansd} uses heatmap-guided losses to enhance native pose awareness. KB-DMGen \cite{liu2025kbdmgen} and Champ \cite{zhu2024champ} integrate global knowledge and 3D geometric information for structural consistency. To improve sparse signal guidance, SP-Ctrl \cite{xuan2025spctrl} and GRPose \cite{yin2025grpose} leverage learnable representations and graph networks to extract topological correlations. Despite single-person success, these methods struggle with multi-person interactions and severe occlusions. Limited by local receptive fields or rudimentary interaction mechanisms, they fail to establish robust long-range dependencies, leading to anatomical distortions. We therefore develop a native pose modeling mechanism within the DiT architecture to eliminate feature interference and overlapping artifacts in multi-subject scenarios.

\begin{figure}[t]
\centering
\includegraphics[width=\linewidth]{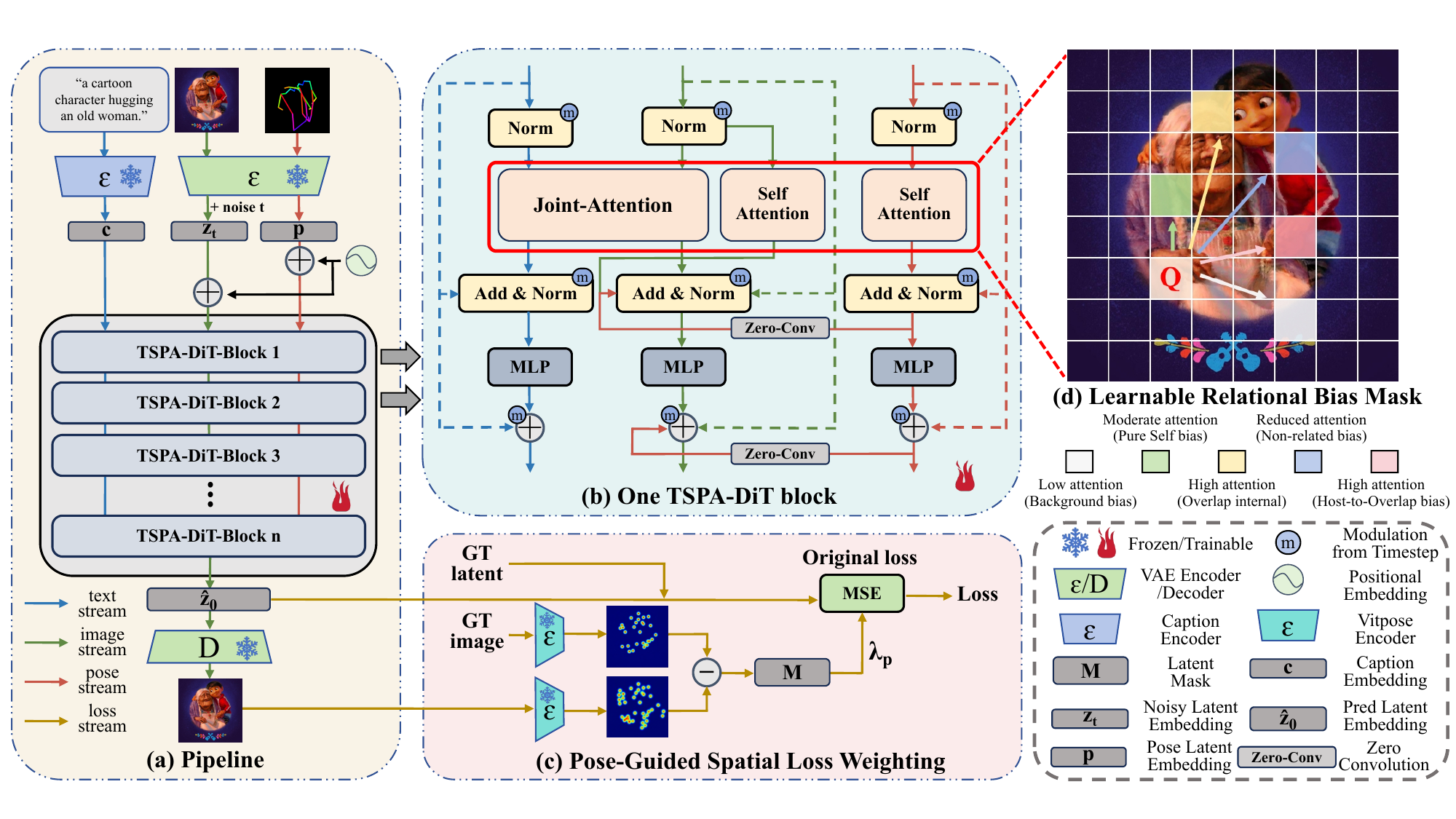}
\caption{\textbf{Overall architecture of TrioPose.} (a) The Triple-Stream Pose-Aware DiT (TSPA-DiT) processes pose as an independent modality. (b) A TSPA-DiT block integrates spatial priors via zero-initialized dual-residual injection to preserve latent stability. (c) Pose-Guided Spatial Loss Weighting modulates the denoising objective using heatmap-derived error maps. (d) The Learnable Relational Bias Mask translates topological connectivity into physical states to decouple inter-instance interference.}
\label{fig:pipeline}
\end{figure}

\section{Method}
\label{sec:method}

\paragraph{Preliminaries: Rectified Flow and MM-DiT}
\label{subsec:preliminaries}

TrioPose integrates skeletal priors into the MM-DiT via a native triple-stream architecture built upon a frozen SD3.5M \cite{esser2024sd3} backbone. To preserve the aligned latent distribution while ensuring efficient, high-precision constrained generation, optimization is restricted to Low-Rank Adaptation (LoRA) \cite{hu2022lora} modules, the novel pose branch, zero-initialized injection layers, and continuous attention biases.

Eschewing discrete noise scheduling, SD3.5M employs continuous-time Rectified Flow (RF) \cite{liu2022rf} to connect data $z_0 \sim p_{\text{data}}$ and noise $\epsilon \sim \mathcal{N}(0, I)$ distributions via a linear trajectory $z_t = t z_0 + (1 - t) \epsilon$ for $t \in [0, 1]$, optimizing a parameterized ODE vector field $v_\theta$ via:
\begin{equation}
    \mathcal{L}_{\text{RF}} = \mathbb{E}_{\epsilon, z_0, t, c} \left[ w(t) \left\| v_\theta(z_t, t, c) - (z_0 - \epsilon) \right\|_2^2 \right],
    \label{eq:rectified_flow_loss}
\end{equation}
where $w(t)$ is an empirical weighting function and $c$ denotes the semantic condition.

Although MM-DiT's dual-stream architecture facilitates semantic alignment by processing text and image modalities independently via joint attention, this isolation complicates the native integration of high-dimensional, dense spatial controls.

\paragraph{Triple-Stream Pose-Aware DiT}
\label{subsec:tri_stream}

To mitigate distribution shift and convergence issues inherent to plug-and-play adapters, we propose a native TSPA-DiT (Fig. \ref{fig:pipeline} (a)) 
. Treating spatially aligned pose latents as an independent modality, a VAE encoder \cite{kingma2013vae} maps them into a 1D sequence $T_{\text{pose}}$ 
. To ensure strict topological alignment, the pose and image streams share identical Spatial Position Embeddings 
. A layer-wise activation mechanism balances geometric constraints with semantic freedom; the Pose Stream activates exclusively in the first $L_p$ layers, reverting to the standard dual-stream backbone thereafter (Sec. \ref{subsec:ablation_study}) 
.

Abstracting layer normalization and signal modulation into pre-processing $\mathcal{F}_{\text{pre}}(\cdot)$ and post-processing $\mathcal{F}_{\text{post}}(\cdot)$ modules, we design a Dual-Residual Zero-Initialized Injection mechanism for smooth prior integration, computing the intermediate pose feature $\tilde{P}^{(l)}$ and output feature $P_{\text{out}}^{(l)}$ as:
\begin{align}
    \tilde{P}^{(l)} &= P_{\text{in}}^{(l)} + \mathcal{F}_{\text{post}}(\text{Attn}(\mathcal{F}_{\text{pre}}(P_{\text{in}}^{(l)}))), \label{eq:pose_middle_coarse} \\
    P_{\text{out}}^{(l)} &= \mathcal{F}_{\text{post}}(\text{MLP}(\mathcal{F}_{\text{pre}}(\tilde{P}^{(l)}))). \label{eq:pose_whole_coarse}
\end{align}
Concurrently, the visual context $I_{\text{out}}^{(l)}$ extracted from the image backbone:
\begin{equation}
    I_{\text{out}}^{(l)} = \mathcal{F}_{\text{post}}(\text{Attn}(\mathcal{F}_{\text{pre}}(X_{\text{in}}^{(l)}))) \label{eq:image_whole_coarse}
\end{equation}
interacts with joint text-image attention, integrating the intermediate pose prior $\tilde{P}^{(l)}$ via a zero-initialized projection ($\text{ZeroConv}_1$):
\begin{equation}
    \tilde{X}^{(l)} = X_{\text{in}}^{(l)} + I_{\text{out}}^{(l)} + \mathcal{F}_{\text{post}}(\text{JointAttn}(\mathcal{F}_{\text{pre}}(X_{\text{in}}^{(l)}), \mathcal{F}_{\text{pre}}(C_{\text{in}}^{(l)}))) + \alpha \cdot \text{ZeroConv}_1(\tilde{P}^{(l)}). \label{eq:backbone_middle_coarse}
\end{equation}
During the subsequent MLP stage, the image feature undergoes non-linear transformation and fuses the final pose topology $P_{\text{out}}^{(l)}$ via a second injection ($\text{ZeroConv}_2$):
\begin{equation}
    X_{\text{out}}^{(l)} = X_{\text{in}}^{(l)} + \mathcal{F}_{\text{post}}(\text{MLP}(\mathcal{F}_{\text{pre}}(\tilde{X}^{(l)}))) + \alpha \cdot \text{ZeroConv}_2(P_{\text{out}}^{(l)}). \label{eq:backbone_whole_coarse}
\end{equation}
This dual-residual structure efficiently injects dense pose priors without compromising pre-trained representations.

\paragraph{Learnable Relational Bias Mask}
\label{subsec:learnable_bias_mask}

\begin{algorithm}[tb]
\caption{Generation of Attention Bias Mask}
\label{alg:bias_mask}
\begin{algorithmic}[1]
    \linespread{1.3}\selectfont
    \State \textbf{Input:} Flattened binarized pose masks $M \in \{0,1\}^{N \times L}$
    \State $C \gets M^\top M$, \quad $T \gets \sum_{n=1}^{N} M[n, :]$, \quad $R \gets \mathbf{0}^{L \times L}$
    \For{$(i, j)$ \textbf{where} $T[i] > 0 \land T[j] > 0$}
        \If{$C[i, j] == 0$} 
            $R[i, j] \gets 3$ \Comment{Non-related}
        \ElsIf{$T[i] == 1 \land T[j] == 1$} 
            $R[i, j] \gets 1$ \Comment{Pure Self}
        \ElsIf{$T[i] \ge 2 \land T[j] \ge 2$} 
            $R[i, j] \gets 2$ \Comment{Overlap Internal}
        \Else 
            ~ $R[i, j] \gets 4$ \Comment{Host-to-Overlap}
        \EndIf
    \EndFor
    \State \textbf{return} $R$
\end{algorithmic}
\end{algorithm}
In multi-subject generation, global self-attention often induces feature crosstalk, whereas hard boolean masks disrupt non-local dependencies. To resolve this, we propose a Learnable Relational Bias Mask that translates discrete topological connectivity into continuous attention biases. First, instance keypoints are rendered as independent 2D skeletal masks via adaptive thickness and Gaussian smoothing. These masks are max-pooled to the latent resolution and binarized into an occupancy matrix $M \in \{0,1\}^{N \times L}$ for $N$ instances and $L$ tokens. We then derive a token occupancy vector $T = \sum_{n=1}^N M[n,:] \in \mathbb{N}^L$ and a co-occurrence matrix $C = M^\top M \in \mathbb{N}^{L \times L}$. Based on $T$ and $C$, token pairs $(i, j)$ are classified into five physical states (Algorithm \ref{alg:bias_mask}): (1) Background ($R=0$), (2) Pure Self ($R=1$), (3) Overlap Internal ($R=2$), (4) Non-related ($R=3$), and (5) Host-to-Overlap ($R=4$).

Discrete relations $R$ are mapped to a continuous bias $B$ to smoothly modulate attention logits:
\begin{equation}
    \text{Attn}(Q, K, V) = \text{Softmax}(QK^\top / \sqrt{d} + B)V.
\label{eq:biased_attention}
\end{equation}
We maintain two sets of five near-zero-initialized learnable scalars: one dedicated to the pose self-attention branch, and another shared by the joint and image self-attention branches. This formulation enables the network to dynamically learn soft relational constraints across occlusion boundaries. To prevent spatial priors from degrading semantic reasoning, we restrict joint bias injection strictly to the image-to-image interaction quadrant. Finally, we employ Bias Mask Dropout \cite{srivastava2014dropout}, randomly zeroing $B$ during training to avert shortcut learning and enhance generative robustness.

\paragraph{Pose-Guided Spatial Loss Weighting}
\label{subsec:spatial_loss_weighting}

Lacking explicit geometric supervision in the Rectified Flow objective, we introduce a Heatmap-Error-Guided Spatial Loss Weighting strategy. Replacing disruptive pose distillation losses, this approach translates local anatomical deviations into continuous latent weight maps. To guarantee global layout stability and prevent large early-stage noise gradients from corrupting the pre-trained latent space, we restrict activation via a Time-step Truncation Constraint:
\begin{equation}
    w_t = \mathbb{I}(t/T \le \tau_t),
\label{eq:timestep_constraint}
\end{equation}
where $\tau_t$ is a predefined threshold.

Under this constraint, a frozen ViTPose \cite{xu2023vitpose} extracts heatmaps ($H^{\text{pred}}$, $H^{\text{gt}}$) from predicted and ground-truth single-person regions. We aggregate the pixel-wise squared error $\mathbf{E} = (H^{\text{pred}} - H^{\text{gt}})^2$ across keypoint channels, binarizing it to isolate severe distortions. This error map is then Gaussian-smoothed and downsampled into a latent mask $M_{\text{latent}}$:
\begin{equation}
    M_{\text{latent}} = \text{Downsample}\big( \text{Gaussian} \big( \bigcup\nolimits_{k=1}^{K} \text{RoI}_k( \mathbb{I}(\mathbf{E}_k > \tau_e) ) \big) \big),
\label{eq:latent_mask}
\end{equation}
where $\tau_e$ defines the error threshold across $K$ valid RoIs.

The base spatial weight is formulated as:
\begin{equation}
    W_{\text{base}} = 1 + \lambda \cdot M_{\text{latent}},
\label{eq:weight_base}
\end{equation}
with $\lambda$ scaling the supervision intensity. Following intra-sample mean normalization to yield $W_{\text{final}}$, this term spatially modulates the original denoising loss $\mathcal{L}_{\text{original}}$:
\begin{equation}
    \mathcal{L}_{\text{total}} = \text{mean}(w_t \cdot W_{\text{final}} \odot \mathcal{L}_{\text{original}}).
\label{eq:weighted_loss}
\end{equation}
Ultimately, this mechanism focuses anatomical supervision precisely on distortion-prone regions while preserving the native diffusion optimization trajectory.

\section{Experiments}
\label{sec:experiment}

\subsection{Experimental Settings}
\label{subsec:experimental_settings}

\paragraph{Datasets.}
\label{subsec:datasets}
TrioPose is evaluated on three benchmark datasets to validate generalization across diverse styles, extreme occlusions, and high-density scenarios. Human-Art\cite{ju2023humanart} serves as the core benchmark for multi-instance decoupling in natural and artistic scenes. OCHuman\cite{zhang2019ochuman} focuses on severe occlusion challenges, while CrowdPose \cite{li2019crowdpose}addresses complex pose modeling in crowded environments.

To leverage the text encoding capabilities of SD3.5-Medium, training annotations are augmented using the Qwen3-VL vision-language model\cite{bai2025qwen3}. A mixed-prompt strategy is designed to accommodate multi-granularity semantic instructions. Specifically, new text labels comprise detailed long descriptions ranging from 80 to 180 words and concise short descriptions ranging from 30 to 60 words at a seven-to-three ratio. This ensures the capture of both fine-grained anatomical attributes and macroscopic environmental semantics. To further enhance robustness in complex multi-person interactions, a mixed training set of 15,500 images is constructed. This set consists of 8,000 randomly sampled Human-Art images, 5,000 CrowdPose images, and all 2,500 OCHuman images. For evaluation, original validation splits are strictly followed to ensure fair comparisons. All human keypoints are preprocessed into the standard COCO-17 format\cite{lin2014coco17}, with relationship matrices generated offline based on instance spatial distributions.

\begin{table}[t]
\centering
\caption{\textbf{Results on Human-Art dataset.} Methods marked with * are evaluated using released checkpoints. \textbf{TrioPose$^\dagger$} denotes the version trained and evaluated with extended long captions. Excluding the foundation baselines (SD* and SD3.5m), the best overall results and the best results among prior competing methods are highlighted in \colorbox{best}{green} and \colorbox{second}{blue}, respectively.}
\label{tab:main_results}
\setlength{\fboxsep}{1.5pt}
\resizebox{\textwidth}{!}{
\begin{tabular}{clcccccc}
\toprule
\multirow{2}{*}{Dataset} & \multirow{2}{*}{Method} & \multicolumn{3}{c}{Pose Accuracy} & \multicolumn{2}{c}{Image Quality} & T2I Alignment \\
\cmidrule(lr){3-5} \cmidrule(lr){6-7} \cmidrule(lr){8-8}
& & AP $\uparrow$ & CAP $\uparrow$ & PCE $\downarrow$ & FID $\downarrow$ & KID $\downarrow$ & CLIP-score $\uparrow$ \\
\midrule
\multirow{9}{*}{Human-Art}
& SD* & 0.24 & 55.71 & 2.30 & 11.53 & 3.36 & 33.33 \\
& SD3.5M & 8.17 & 80.57 & 0.92 & 2.17 & 1.66 & 34.50 \\
\cmidrule{2-8}
& T2I-Adapter & 27.22 & 65.65 & 1.75 & 11.92 & 2.73 & \colorbox{second}{33.27} \\
& ControlNet & 39.52 & 69.19 & 1.54 & 11.01 & \colorbox{second}{2.23} & 32.65 \\
& Uni-ControlNet & 41.94 & 69.32 & 1.48 & 14.63 & 2.30 & 32.51 \\
& GLIGEN & 18.24 & 69.15 & 1.46 & -- & -- & 32.52 \\
& HumanSD & 44.57 & 69.68 & \colorbox{second}{1.37} & \colorbox{second}{10.03} & 2.70 & 32.24 \\
& Stable-Pose & 48.88 & 70.83 & 1.50 & 11.12 & 2.35 & 32.60 \\
& GRPose & \colorbox{second}{49.50} & \colorbox{second}{70.84} & 1.43 & 13.76 & 2.53 & 32.31 \\
\cmidrule{2-8}
& TrioPose(Ours) & 59.24 & 82.94 & 0.91 & 3.03 & 1.22 & 33.50 \\
& TrioPose$^\dagger$(Ours) & \colorbox{best}{64.33} & \colorbox{best}{83.23} & \colorbox{best}{0.86} & \colorbox{best}{1.65} & \colorbox{best}{0.87} & \colorbox{best}{34.46} \\
\bottomrule
\end{tabular}
}
\end{table}

\subsection{Comparison with SOTA Methods}
\label{subsec:comparison_with_sota}

\paragraph{Quantitative Evaluation.}
Table \ref{tab:main_results} reports performance on the Human-Art dataset using AP, CAP \cite{posenetsimilarity}, PCE \cite{cheong2022pce}, FID \cite{heusel2017fid}, KID \cite{binkowski2018kid} (scaled by 100), and CLIP-score \cite{radford2021clip-score}. Using original annotations, TrioPose achieves $59.24$ AP and $82.94$ CAP, outperforming the previous, GRPose \cite{yin2025grpose}, by approximately $20\%$. This demonstrates the triple-stream architecture's superiority in precise pose alignment under identical constraints. 

With the mixed-prompt strategy, TrioPose$^\dagger$ leverages extended captions to engage SD3.5M's deep semantic priors, elevating AP to a state-of-the-art $64.33$. For image quality, TrioPose$^\dagger$ yields the lowest FID ($1.65$), significantly surpassing HumanSD ($10.03$) \cite{ju2023humansd} and Stable-Pose ($11.12$) \cite{wang2024stable} , successfully avoiding the severe texture loss common in traditional feature concatenation methods. 

Additionally, TrioPose$^\dagger$'s peak CLIP-score ($34.46$) confirms the framework strengthens spatial constraints without compromising semantic alignment. Overall, these results highlight TrioPose's robust pose control and visual fidelity, establishing a new SOTA standard for complex human generation.

\begin{figure}[t]
\centering
\includegraphics[width=\linewidth]{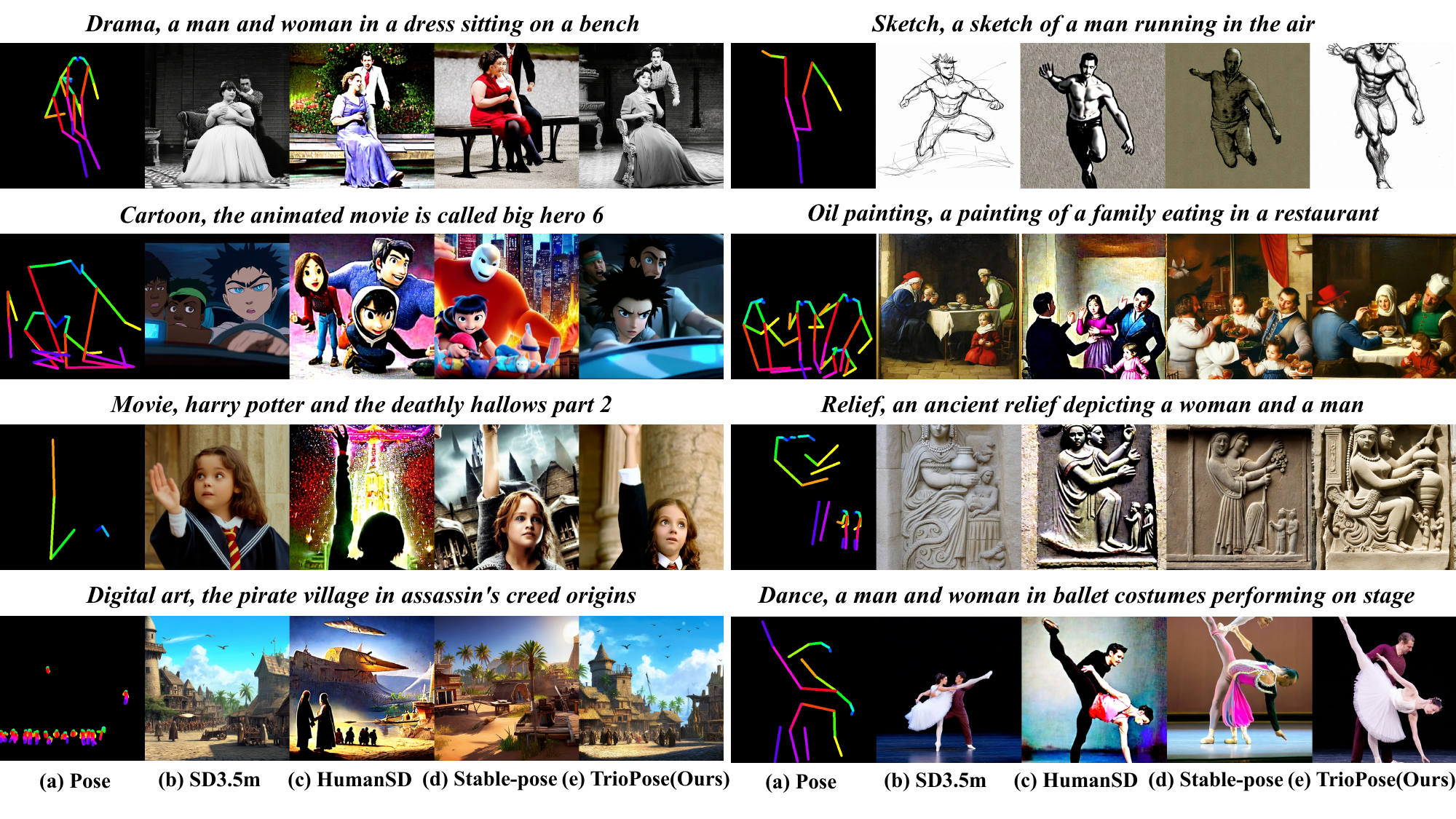}
\caption{\textbf{Qualitative comparisons.} Compared to SD3.5M, HumanSD, and Stable-Pose across diverse styles and complex occlusions, TrioPose achieves precise anatomical alignment and robust instance decoupling, effectively eliminating the feature crosstalk prevalent in baselines.}
\label{fig:Qualitative Comparison}
\end{figure}

\paragraph{Qualitative Comparison.}
Qualitative analysis demonstrates TrioPose's advantages in anatomical rigor and spatial consistency. In complex scenarios with entangled limbs (e.g., Drama and Cartoon in Fig. \ref{fig:Qualitative Comparison}), TrioPose exhibits exceptional instance decoupling. Modulated by the learnable relational bias mask, attention flows maintain topological integrity in distortion-prone regions like hands and legs. Conversely, HumanSD and Stable-Pose frequently suffer from feature entanglement during severe overlaps, such as incorrectly attaching limb A to body B. 

TrioPose remains robust even under imperfect pose guidance. Despite missing limbs in the input pose image (e.g., Sketch, Movie, and Relief in Fig. \ref{fig:Qualitative Comparison}), it reconstructs absent anatomical details by combining triple-stream structural modeling with global semantic priors. This self-consistency ensures physical logic and prevents inheriting input defects. 

In scenes with numerous small figures (e.g., Digital art in Fig. \ref{fig:Qualitative Comparison}), TrioPose maintains precise micro-scale pose alignment and clear boundaries, validating the native diffusion transformer's capability in cross-scale spatial control. This visual purity, achieved via the synergy of relational biases and spatial loss weighting, effectively eliminates signal diffusion into irrelevant regions.

\paragraph{Robustness in Complex Multi-Human Scenarios}
\label{subsec:robustness_complex_scenarios}

To validate generalization in high-density and occluded scenes, cross-dataset evaluations are performed using the mixed training configuration. As shown in Table \ref{tab:mixed_dataset_results}, TrioPose consistently outperforms Stable-Pose in pose and semantic alignment across three distributionally diverse independent validation sets.

On the multi-style Human-Art validation set, TrioPose achieves $61.52$ AP and $1.93$ FID, compared to Stable-Pose's $58.24$ and $4.30$. This substantial margin confirms the native triple-stream architecture's generalization in parsing complex semantics and diverse stylistic mappings. 
\setlength{\columnsep}{15pt}
\begin{wraptable}{r}{0.55\textwidth}
    \centering
    \vspace{-5pt}
    \caption{\textbf{Evaluation on individual validation sets after mixed-dataset training.} Both Stable-Pose and TrioPose are trained on the 15.5k mixed dataset with Qwen3-VL long captions. The best results for each dataset are highlighted in bold.}
    \label{tab:mixed_dataset_results}
    \vspace{10pt}
    \resizebox{\linewidth}{!}{
    \begin{tabular}{llccc}
        \toprule
        Dataset & Method & AP $\uparrow$ & FID $\downarrow$ & CLIP $\uparrow$ \\
        \midrule
        & Stable-Pose & 58.24 & 4.30 & 32.17 \\
        \rowcolor{gray!15} \cellcolor{white}\multirow{-2}{*}{Human-Art} 
        & TrioPose (Ours) & \textbf{61.52} & \textbf{1.93} & \textbf{34.48} \\
        \midrule
        
        & Stable-Pose & 50.25 & \textbf{0.65} & 30.98 \\
        \rowcolor{gray!15} \cellcolor{white}\multirow{-2}{*}{CrowdPose} 
        & TrioPose (Ours) & \textbf{58.56} & 0.78 & \textbf{32.85} \\
        \midrule
        
        & Stable-Pose & 58.84 & \textbf{0.89} & 30.86 \\
        \rowcolor{gray!15} \cellcolor{white}\multirow{-2}{*}{OCHuman} 
        & TrioPose (Ours) & \textbf{62.59} & 0.92 & \textbf{31.86} \\
        \bottomrule
    \end{tabular}
    }
    \vspace{-10pt}
\end{wraptable}
For high-density CrowdPose scenarios, AP significantly improves from $50.25$ to $58.56$. Although the $0.78$ FID slightly trails Stable-Pose's $0.65$, this minor variance preserves visual quality while yielding substantial pose accuracy gains. On the highly occluded OCHuman set, TrioPose achieves $62.59$ AP. The marginal FID difference ($0.92$ vs. $0.89$) indicates enhanced pose control under extreme occlusion without visual degradation.

These cross-dataset results reflect robust generalization, demonstrating effective extraction of structural priors from the mixed training set. Furthermore, TrioPose achieves higher CLIP scores across all sets, proving the text augmentation strategy ensures stable text-to-image semantic alignment. Overall, TrioPose optimally balances spatial constraints, visual fidelity, and semantic consistency in complex multi-person scenarios.
\subsection{Ablation Study}
\label{subsec:ablation_study}

\begin{table}[h]
\centering
\caption{Ablation study on the effectiveness of proposed modules. The Triple-Stream Pose-Aware DiT (TSPA), Learnable Relational Bias Mask (LRBM), and Pose-Guided Spatial Loss Weighting (PSLW) are progressively integrated into the baseline model.$\circ$ denotes the substitution of our proposed LRBM with a traditional binary hard mask. The best results for each metric are highlighted in bold.}
\label{tab:ablation}
\begin{tabular}{ccccccccc}
\toprule
\multicolumn{3}{c}{Modules} & \multicolumn{3}{c}{Pose Accuracy} & \multicolumn{2}{c}{Image Quality} & T2I Alignment \\
\cmidrule(lr){1-3} \cmidrule(lr){4-6} \cmidrule(lr){7-8} \cmidrule(lr){9-9}
TSPA & LRBM & PSLW & AP $\uparrow$ & CAP $\uparrow$ & PCE $\downarrow$ & FID $\downarrow$ & KID $\downarrow$ & CLIP-score $\uparrow$ \\
\midrule
 $\times$   & $\times$   & $\times$   & 8.17  & 80.57 & 0.92 & 2.17 & 1.66 & 34.50 \\
 \checkmark & $\times$   & $\times$   & 63.50 & \textbf{83.32} & \textbf{0.85} & 2.03 & 1.31 & 34.39 \\
 \checkmark & \checkmark & $\times$   & \textbf{64.34} & 83.25 & 0.86 & 1.95 & 1.17 & 34.39 \\
 \checkmark & $\times$   & \checkmark & 63.37 & 83.16 & 0.88 & 1.82 & 0.98 & 34.45 \\
 \checkmark & $\circ$    & \checkmark & 62.63 & 83.14 & 0.87 & 1.82 & 1.03 & \textbf{34.58} \\
 \rowcolor{gray!15} 
 \checkmark & \checkmark & \checkmark & 64.33 & 83.23 & 0.86 & \textbf{1.65} & \textbf{0.87} & 34.46 \\
\bottomrule
\end{tabular}
\end{table}

\paragraph{Component-wise Efficacy.}
Progressive ablation studies, shown in Table \ref{tab:ablation}, validate the proposed modules. Transitioning to the TSPA architecture alone yields a substantial AP increase from $8.17$ to $63.50$, confirming that native pose modeling within the DiT framework\cite{peebles2023dit} resolves structural collapse during pose-guided generation. 

Incorporating LRBM further improves AP to $64.34$ in multi-person scenarios, proving soft-constrained attention modulation across physical relations effectively decouples cross-instance feature entanglement. Furthermore, replacing our fine-grained 5-state relational mask with a traditional binary hard mask degrades AP to $62.63$ and increases FID to $1.82$, proving that nuanced topological division is essential for resolving complex limb overlaps rather than simple boolean isolation. Integrating PSLW stabilizes AP gains while significantly optimizing visual quality, reducing FID and KID to $1.65$ and $0.87$, respectively. Modulating the native denoising objective with spatial weight maps corrects micro-anatomical distortions without compromising fidelity. Ultimately, synergizing these three core mechanisms optimally balances robust pose constraints and high-fidelity image synthesis.

\paragraph{Optimization of TSPA-DiT Configuration.}
Sensitivity analysis of the triple-stream architecture focuses on pose stream activation layers ($L_p$) and injection scale ($\alpha$). As shown in Table \ref{tab:ts_pa_dit_ablation}, $L_p=12$ optimally balances geometric constraints and image quality, yielding the highest AP ($64.33$) and lowest FID ($1.65$). This establishes a spatial geometric framework in lower layers while preserving semantic freedom in higher layers. Shallow activation ($L_p=6$) provides inadequate constraints for complex distortions, whereas deep activation ($L_p=24$) imposes rigid spatial suppression, degrading diversity and increasing FID to $2.27$.

Adjusting $\alpha$ reveals a similar pattern, where $\alpha=0.5$ yields optimal stability and alignment. Coupled with zero-initialized \cite{zhang2023controlnet} dual-residual injection, this ensures robust pose guidance without disrupting the pre-trained latent distribution\cite{rombach2022sd1.5}. Lower intensity ($\alpha=0.3$) provides insufficient guidance, while excessive injection ($\alpha=0.7$) interferes with the latent space, degrading both AP and FID. This synergy of layer-wise activation and controlled scaling forms the architectural foundation for robust pose awareness.
\begin{table}[htbp]
    \centering
    \caption{\textbf{Ablation study on TS-PA-DiT configuration.} Top: ablating the number of activated layers ($L_p$). Bottom: ablating the injection scale factor ($\alpha$). The default setting is highlighted in gray.}
    \label{tab:ts_pa_dit_ablation}
    \setlength{\tabcolsep}{8pt}
    \begin{tabular}{@{}cc cccccc@{}}
        \toprule
        \multicolumn{2}{c}{Config} & \multicolumn{3}{c}{Pose Accuracy} & \multicolumn{2}{c}{Image Quality} & T2I Alignment \\
        \cmidrule(lr){1-2} \cmidrule(lr){3-5} \cmidrule(lr){6-7} \cmidrule(lr){8-8}
        $L_p$ & $\alpha$ & AP $\uparrow$ & CAP $\uparrow$ & PCE $\downarrow$ & FID $\downarrow$ & KID $\downarrow$ & CLIP $\uparrow$ \\
        \midrule
        6  & 0.5 & 63.03 & \textbf{83.34} & 0.88 & 1.78 & 1.03 & \textbf{34.49} \\
        \rowcolor{gray!15} 
        12 & 0.5 & \textbf{64.33} & 83.23 & 0.86 & \textbf{1.65} & \textbf{0.87} & 34.46 \\
        24 & 0.5 & 62.70 & 82.97 & \textbf{0.85} & 2.27 & 1.17 & 34.37 \\
        \midrule
        12 & 0.3 & 64.19 & 83.25 & 0.87 & 2.00 & 0.98 & 34.43 \\
        \rowcolor{gray!15} 
        12 & 0.5 & \textbf{64.33} & 83.23 & 0.86 & \textbf{1.65} & \textbf{0.87} & \textbf{34.46} \\
        12 & 0.7 & 62.87 & \textbf{83.20} & \textbf{0.85} & 2.01 & 0.93 & 34.37 \\
        \bottomrule
    \end{tabular}
\end{table}

\begin{table}[htbp]
    \centering
    \caption{\textbf{Ablation study on relational bias initialization.} We compare fixed (non-learnable) biases with learnable biases under different initialization strategies. By employing empirical topological priors ($\boldsymbol{\theta}_{\text{prior}}$) derived from the steady-state convergence trajectories of zero-initialized parameters, the proposed model effectively circumvents early-stage gradient oscillation and achieves superior spatial guidance compared to the zero initialization ($\mathbf{0}$) baseline.}
    \label{tab:ablation_bias_init}
    \setlength{\tabcolsep}{7pt}
    \begin{tabular}{@{}l cccccc@{}}
        \toprule
        & \multicolumn{3}{c}{Pose Accuracy} & \multicolumn{2}{c}{Image Quality} & T2I Alignment \\
        \cmidrule(lr){2-4} \cmidrule(lr){5-6} \cmidrule(lr){7-7}
        Bias Strategy & AP $\uparrow$ & CAP $\uparrow$ & PCE $\downarrow$ & FID $\downarrow$ & KID $\downarrow$ & CLIP-score $\uparrow$ \\
        \midrule
        Fixed Bias & 62.93 & 83.16 & 0.87 & 2.08 & 1.02 & 34.43 \\
        Learnable ($\boldsymbol{\theta} = \mathbf{0}$) & 63.12 & \textbf{83.30} & 0.88 & 1.81 & 0.96 & 34.45 \\
        \rowcolor{gray!15} 
        Learnable ($\boldsymbol{\theta} = \boldsymbol{\theta}_{\text{prior}}$) & \textbf{64.33} & 83.23 & \textbf{0.86} & \textbf{1.65} & \textbf{0.87} & \textbf{34.46} \\
        \bottomrule
    \end{tabular}
\end{table}

\paragraph{Initialization Strategies for Relational Bias Mask.}
\label{subsec: initialization strategies for relational bias mask.}

Table \ref{tab:ablation_bias_init} compares relational bias mask initialization strategies: fixed non-learnable biases, zero initialization ($\boldsymbol{\theta} = \mathbf{0}$), and physics-derived prior-guided initialization ($\boldsymbol{\theta}_{\text{prior}}$). While learnable parameters effectively lower FID and KID compared to fixed biases, uniform zero initialization restricts pose alignment accuracy, yielding only $63.12$ AP. 

Training dynamics reveal that zero-initialized parameters naturally converge toward specific physical magnitudes, establishing the topological prior $\boldsymbol{\theta}_{\text{prior}}$. Specifically, associated states (e.g., pure self, overlap internal) receive positive initializations, while physically inhibitory connections (non-related subjects) are uniformly initialized to $-0.3$. This physics-guided initialization substantially increases AP to an optimal $64.33$ and minimizes FID to $1.65$. Ultimately, this validates that rational topological priors provide precise spatial guidance, enabling multimodal attention layers to bypass early-stage gradient oscillations and smoothly converge toward an optimal feature space.

\setlength{\columnsep}{5pt} 

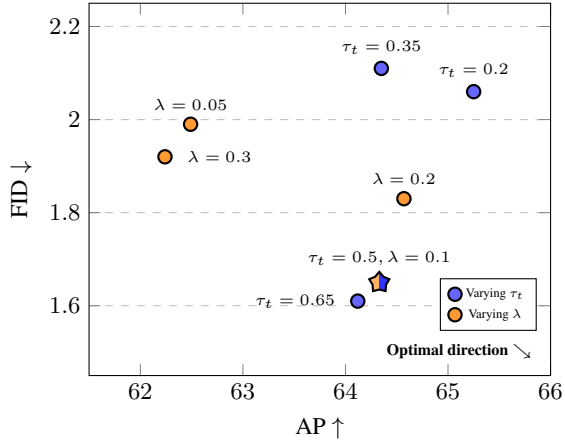
\begin{wrapfigure}[18]{r}{0.55\textwidth}
    
    \vspace{-10pt} 

    \centering
    \begin{tikzpicture}
        \begin{axis}[
            width=\linewidth,
            height=6.5cm,
            xlabel={AP $\uparrow$},
            ylabel={FID $\downarrow$},
            xmin=61.5, xmax=66,
            ymin=1.45, ymax=2.25,
            ymajorgrids=true,
            grid style=dashed,
            legend style={
                font=\tiny, 
                nodes={scale=0.8, transform shape},
                at={(0.97, 0.12)}, 
                anchor=south east
            }, 
            tick label style={font=\small, inner sep=3pt},
            label style={font=\small},
        ]
        
        \addplot[
            only marks,
            mark=*,
            mark size=2.5pt,
            draw=black,
            fill=blue!60!white,
            thick
        ] coordinates {
            (65.25, 2.06) 
            (64.35, 2.11) 
            (64.12, 1.61) 
        };
        \addlegendentry{Varying $\tau_t$}
        
        \addplot[
            only marks,
            mark=*,
            mark size=2.5pt,
            draw=black,
            fill=orange!80!white,
            thick
        ] coordinates {
            (62.24, 1.92) 
            (62.49, 1.99) 
            (64.57, 1.83) 
        };
        \addlegendentry{Varying $\lambda$}
        
        \node[
            star,           
            draw=black,     
            thick,          
            minimum size=8pt, 
            inner sep=0pt,
            path picture={
                \fill[orange!60!white] (path picture bounding box.south west) rectangle (path picture bounding box.north);
                \fill[blue!80!white] (path picture bounding box.south) rectangle (path picture bounding box.north east);
            }
        ] at (axis cs:64.33, 1.65) {};
        
        \node[above=2pt, font=\tiny] at (axis cs:65.25, 2.06) {$\tau_t=0.2$};
        \node[above=2pt, font=\tiny] at (axis cs:64.35, 2.11) {$\tau_t=0.35$};
        \node[left=5pt, font=\tiny] at (axis cs:64.12, 1.61) {$\tau_t=0.65$};
        \node[right=5pt, font=\tiny] at (axis cs:62.24, 1.92) {$\lambda=0.3$};
        \node[above=2pt, font=\tiny] at (axis cs:64.57, 1.83) {$\lambda=0.2$};
        \node[above=2pt, font=\tiny] at (axis cs:62.49, 1.99) {$\lambda=0.05$};
        
        \node[above=3pt, font=\tiny] at (axis cs:64.33, 1.65) {$\tau_t=0.5, \lambda=0.1$};
        \node[anchor=south east, font=\tiny] at (rel axis cs:0.98, 0.02) {\textbf{Optimal direction} $\searrow$};
        \end{axis}
    \end{tikzpicture}
    
    \vspace{-10pt} 
    
    \caption{Impact of hyper-parameters $\tau_t$ and $\lambda$ on the AP-FID trade-off.}
    \label{fig:loss_ablation}
    
    \vspace{-15pt} 

\end{wrapfigure}

\setlength{\columnsep}{18pt} 

\paragraph{Impact of Spatial Loss Weighting.}
Sensitivity analysis on the truncation threshold $\tau_t$ (Eq. \ref{eq:timestep_constraint}) and spatial weight coefficient $\lambda$ (Eq. \ref{eq:weight_base}) investigates the trade-off between anatomical supervision (AP) and generative fidelity (FID). As shown in Fig. \ref{fig:loss_ablation}, the optimal performance direction is the bottom-right. For $\tau_t$, smaller values ($0.2$ or $0.35$) cause a significant FID surge, indicating early-stage intense guidance severely disrupts visual fidelity despite marginal pose accuracy gains. Increasing $\tau_t$ to $0.5$ shifts performance optimally, drastically improving image quality while maintaining precise alignment.

Similarly, $\lambda$ is critical. An excessively strong weight ($\lambda=0.3$) collapses both AP and FID, proving that overpowering spatial constraints disrupt the diffusion model's native denoising logic. Evaluating the distribution trajectories, $\tau_t=0.5$ and $\lambda=0.1$ (dual-colored star in Fig. \ref{fig:loss_ablation}) is selected as the optimal configuration, effectively balancing robust spatial constraints with natural visual synthesis.

\section{Conclusion}
\label{sec:conclusion}

We propose TrioPose, a native pose-driven framework that addresses feature crosstalk and anatomical distortion in multi-person image generation. By treating pose as an independent modality within the TSPA-DiT architecture and employing a Learnable Relational Bias Mask, our method effectively decouples complex instance interactions while preserving the pre-trained MM-DiT distribution. Extensive evaluations demonstrate that TrioPose establishes a new state-of-the-art across Human-Art, CrowdPose, and OCHuman benchmarks, achieving a superior balance between spatial control and visual fidelity.

However, TrioPose's high-fidelity synthesis capabilities introduce potential misuse risks, notably Deepfakes. To mitigate these societal impacts and ensure responsible deployment, we will integrate digital watermarking into our released checkpoints to facilitate robust provenance tracking. Future work will explore lightweight attention mechanisms to accelerate inference and extend the framework for temporally consistent pose-guided video synthesis.

\bibliographystyle{plain}
\bibliography{main.bib}

@misc{posenetsimilarity,
  author = {freshsomebody},
  title = {Posenet similarity},
  year = {2023},
  publisher = {GitHub},
  journal = {GitHub repository},
  howpublished = {\url{https://github.com/freshsomebody/posenet-similarity}},
  note = {Accessed: 2026-05-05}
}

@inproceedings{rombach2022sd1.5,
  title={High-resolution image synthesis with latent diffusion models},
  author={Rombach, Robin and Blattmann, Andreas and Lorenz, Dominik and Esser, Patrick and Ommer, Bj{\"o}rn},
  booktitle={Proceedings of the IEEE/CVF conference on computer vision and pattern recognition},
  pages={10684--10695},
  year={2022}
}

@inproceedings{esser2024sd3,
  title={Scaling rectified flow transformers for high-resolution image synthesis},
  author={Esser, Patrick and Kulal, Sumith and Blattmann, Andreas and Entezari, Rahim and M{\"u}ller, Jonas and Saini, Harry and Levi, Yam and Lorenz, Dominik and Sauer, Axel and Boesel, Frederic and others},
  booktitle={Forty-first international conference on machine learning},
  year={2024}
}

@article{zhao2023unicontrol,
  title={Uni-controlnet: All-in-one control to text-to-image diffusion models},
  author={Zhao, Shihao and Chen, Dongdong and Chen, Yen-Chun and Bao, Jianmin and Hao, Shaozhe and Yuan, Lu and Wong, Kwan-Yee K},
  journal={Advances in neural information processing systems},
  volume={36},
  pages={11127--11150},
  year={2023}
}

@inproceedings{li2023gligen,
  title={Gligen: Open-set grounded text-to-image generation},
  author={Li, Yuheng and Liu, Haotian and Wu, Qingyang and Mu, Fangzhou and Yang, Jianwei and Gao, Jianfeng and Li, Chunyuan and Lee, Yong Jae},
  booktitle={Proceedings of the IEEE/CVF conference on computer vision and pattern recognition},
  pages={22511--22521},
  year={2023}
}

@inproceedings{ju2023humansd,
  title={Humansd: A native skeleton-guided diffusion model for human image generation},
  author={Ju, Xuan and Zeng, Ailing and Zhao, Chenchen and Wang, Jianan and Zhang, Lei and Xu, Qiang},
  booktitle={Proceedings of the IEEE/CVF International Conference on Computer Vision},
  pages={15988--15998},
  year={2023}
}

@article{wang2024stable,
  title={Stable-pose: Leveraging transformers for pose-guided text-to-image generation},
  author={Wang, Jiajun and Ghahremani, Morteza and Li, Yitong and Ommer, Bj{\"o}rn and Wachinger, Christian},
  journal={Advances in Neural Information Processing Systems},
  volume={37},
  pages={65670--65698},
  year={2024}
}

@inproceedings{mou2024t2i,
  title={T2i-adapter: Learning adapters to dig out more controllable ability for text-to-image diffusion models},
  author={Mou, Chong and Wang, Xintao and Xie, Liangbin and Wu, Yanze and Zhang, Jian and Qi, Zhongang and Shan, Ying},
  booktitle={Proceedings of the AAAI conference on artificial intelligence},
  volume={38},
  number={5},
  pages={4296--4304},
  year={2024}
}

@inproceedings{zhang2023controlnet,
  title={Adding conditional control to text-to-image diffusion models},
  author={Zhang, Lvmin and Rao, Anyi and Agrawala, Maneesh},
  booktitle={Proceedings of the IEEE/CVF international conference on computer vision},
  pages={3836--3847},
  year={2023}
}

@inproceedings{xuan2025spctrl,
  title={Rethink Sparse Signals for Pose-guided Text-to-image Generation},
  author={Xuan, Wenjie and Zhang, Jing and Liu, Juhua and Du, Bo and Tao, Dacheng},
  booktitle={Proceedings of the IEEE/CVF International Conference on Computer Vision},
  pages={15896--15906},
  year={2025}
}

@inproceedings{yin2025grpose,
  title={Grpose: Learning graph relations for human image generation with pose priors},
  author={Yin, Xiangchen and Di, Donglin and Fan, Lei and Li, Hao and Chen, Wei and Song, Yang and Sun, Xiao and Yang, Xun and others},
  booktitle={Proceedings of the AAAI Conference on Artificial Intelligence},
  volume={39},
  number={9},
  pages={9526--9534},
  year={2025}
}

@article{xu2023vitpose,
  title={Vitpose++: Vision transformer for generic body pose estimation},
  author={Xu, Yufei and Zhang, Jing and Zhang, Qiming and Tao, Dacheng},
  journal={IEEE Transactions on Pattern Analysis and Machine Intelligence},
  volume={46},
  number={2},
  pages={1212--1230},
  year={2023},
  publisher={IEEE}
}

@inproceedings{ju2023humanart,
  title={Human-art: A versatile human-centric dataset bridging natural and artificial scenes},
  author={Ju, Xuan and Zeng, Ailing and Wang, Jianan and Xu, Qiang and Zhang, Lei},
  booktitle={Proceedings of the IEEE/CVF conference on computer vision and pattern recognition},
  pages={618--629},
  year={2023}
}

@article{pan2025pixelponder,
  title={Pixelponder: Dynamic patch adaptation for enhanced multi-conditional text-to-image generation},
  author={Pan, Yanjie and He, Qingdong and Jiang, Zhengkai and Xu, Pengcheng and Wang, Chaoyi and Peng, Jinlong and Wang, Haoxuan and Cao, Yun and Gan, Zhenye and Chi, Mingmin and others},
  journal={arXiv preprint arXiv:2503.06684},
  year={2025}
}

@article{ma2017pg2,
  title={Pose guided person image generation},
  author={Ma, Liqian and Jia, Xu and Sun, Qianru and Schiele, Bernt and Tuytelaars, Tinne and Van Gool, Luc},
  journal={Advances in neural information processing systems},
  volume={30},
  year={2017}
}

@inproceedings{zhu2019patn,
  title={Progressive pose attention transfer for person image generation},
  author={Zhu, Zhen and Huang, Tengteng and Shi, Baoguang and Yu, Miao and Wang, Bofei and Bai, Xiang},
  booktitle={Proceedings of the IEEE/CVF conference on computer vision and pattern recognition},
  pages={2347--2356},
  year={2019}
}

@inproceedings{grigorev2019coordinate,
  title={Coordinate-based texture inpainting for pose-guided human image generation},
  author={Grigorev, Artur and Sevastopolsky, Artem and Vakhitov, Alexander and Lempitsky, Victor},
  booktitle={Proceedings of the IEEE/CVF Conference on Computer Vision and Pattern Recognition},
  pages={12135--12144},
  year={2019}
}

@inproceedings{zhang2021pise,
  title={Pise: Person image synthesis and editing with decoupled gan},
  author={Zhang, Jinsong and Li, Kun and Lai, Yu-Kun and Yang, Jingyu},
  booktitle={Proceedings of the IEEE/CVF conference on computer vision and pattern recognition},
  pages={7982--7990},
  year={2021}
}

@article{liu2025kbdmgen,
  title={KB-DMGen: Knowledge-Based Global Guidance and Dynamic Pose Masking for Human Image Generation},
  author={Liu, Shibang and Xie, Xuemei and Shi, Guangming},
  journal={arXiv preprint arXiv:2507.20083},
  year={2025}
}

@inproceedings{zhu2024champ,
  title={Champ: Controllable and consistent human image animation with 3d parametric guidance},
  author={Zhu, Shenhao and Chen, Junming Leo and Dai, Zuozhuo and Dong, Zilong and Xu, Yinghui and Cao, Xun and Yao, Yao and Zhu, Hao and Zhu, Siyu},
  booktitle={European Conference on Computer Vision},
  pages={145--162},
  year={2024},
  organization={Springer}
}

@article{goodfellow2014gan,
  title={Generative adversarial nets},
  author={Goodfellow, Ian J and Pouget-Abadie, Jean and Mirza, Mehdi and Xu, Bing and Warde-Farley, David and Ozair, Sherjil and Courville, Aaron and Bengio, Yoshua},
  journal={Advances in neural information processing systems},
  volume={27},
  year={2014}
}

@article{hu2022lora,
  title={Lora: Low-rank adaptation of large language models.},
  author={Hu, Edward J and Shen, Yelong and Wallis, Phillip and Allen-Zhu, Zeyuan and Li, Yuanzhi and Wang, Shean and Wang, Liang and Chen, Weizhu and others},
  journal={Iclr},
  volume={1},
  number={2},
  pages={3},
  year={2022}
}

@article{liu2022rf,
  title={Flow straight and fast: Learning to generate and transfer data with rectified flow},
  author={Liu, Xingchao and Gong, Chengyue and Liu, Qiang},
  journal={arXiv preprint arXiv:2209.03003},
  year={2022}
}

@article{kingma2013vae,
  title={Auto-encoding variational bayes},
  author={Kingma, Diederik P and Welling, Max},
  journal={arXiv preprint arXiv:1312.6114},
  year={2013}
}

@inproceedings{li2019crowdpose,
  title={Crowdpose: Efficient crowded scenes pose estimation and a new benchmark},
  author={Li, Jiefeng and Wang, Can and Zhu, Hao and Mao, Yihuan and Fang, Hao-Shu and Lu, Cewu},
  booktitle={Proceedings of the IEEE/CVF conference on computer vision and pattern recognition},
  pages={10863--10872},
  year={2019}
}

@inproceedings{zhang2019ochuman,
  title={Pose2seg: Detection free human instance segmentation},
  author={Zhang, Song-Hai and Li, Ruilong and Dong, Xin and Rosin, Paul and Cai, Zixi and Han, Xi and Yang, Dingcheng and Huang, Haozhi and Hu, Shi-Min},
  booktitle={Proceedings of the IEEE/CVF conference on computer vision and pattern recognition},
  pages={889--898},
  year={2019}
}

@article{bai2025qwen3,
  title={Qwen3-vl technical report},
  author={Bai, Shuai and Cai, Yuxuan and Chen, Ruizhe and Chen, Keqin and Chen, Xionghui and Cheng, Zesen and Deng, Lianghao and Ding, Wei and Gao, Chang and Ge, Chunjiang and others},
  journal={arXiv preprint arXiv:2511.21631},
  year={2025}
}

@inproceedings{lin2014coco17,
  title={Microsoft coco: Common objects in context},
  author={Lin, Tsung-Yi and Maire, Michael and Belongie, Serge and Hays, James and Perona, Pietro and Ramanan, Deva and Doll{\'a}r, Piotr and Zitnick, C Lawrence},
  booktitle={European conference on computer vision},
  pages={740--755},
  year={2014},
  organization={Springer}
}

@article{heusel2017fid,
  title={Gans trained by a two time-scale update rule converge to a local nash equilibrium},
  author={Heusel, Martin and Ramsauer, Hubert and Unterthiner, Thomas and Nessler, Bernhard and Hochreiter, Sepp},
  journal={Advances in neural information processing systems},
  volume={30},
  year={2017}
}

@article{binkowski2018kid,
  title={Demystifying mmd gans},
  author={Bi{\'n}kowski, Miko{\l}aj and Sutherland, Danica J and Arbel, Michael and Gretton, Arthur},
  journal={arXiv preprint arXiv:1801.01401},
  year={2018}
}

@inproceedings{radford2021clip-score,
  title={Learning transferable visual models from natural language supervision},
  author={Radford, Alec and Kim, Jong Wook and Hallacy, Chris and Ramesh, Aditya and Goh, Gabriel and Agarwal, Sandhini and Sastry, Girish and Askell, Amanda and Mishkin, Pamela and Clark, Jack and others},
  booktitle={International conference on machine learning},
  pages={8748--8763},
  year={2021},
  organization={PmLR}
}

@inproceedings{peebles2023dit,
  title={Scalable diffusion models with transformers},
  author={Peebles, William and Xie, Saining},
  booktitle={Proceedings of the IEEE/CVF international conference on computer vision},
  pages={4195--4205},
  year={2023}
}

@article{cheong2022pce,
  title={Kpe: Keypoint pose encoding for transformer-based image generation},
  author={Cheong, Soon Yau and Mustafa, Armin and Gilbert, Andrew},
  journal={arXiv preprint arXiv:2203.04907},
  year={2022}
}

@article{srivastava2014dropout,
  title={Dropout: a simple way to prevent neural networks from overfitting},
  author={Srivastava, Nitish and Hinton, Geoffrey and Krizhevsky, Alex and Sutskever, Ilya and Salakhutdinov, Ruslan},
  journal={The journal of machine learning research},
  volume={15},
  number={1},
  pages={1929--1958},
  year={2014},
  publisher={JMLR. org}
}

@article{ho2020ddpm,
  title={Denoising diffusion probabilistic models},
  author={Ho, Jonathan and Jain, Ajay and Abbeel, Pieter},
  journal={Advances in neural information processing systems},
  volume={33},
  pages={6840--6851},
  year={2020}
}

@article{liu2022verbal,
  title={Verbal-person nets: Pose-guided multi-granularity language-to-person generation},
  author={Liu, Deyin and Wu, Lin and Zheng, Feng and Liu, Lingqiao and Wang, Meng},
  journal={IEEE Transactions on Neural Networks and Learning Systems},
  volume={34},
  number={11},
  pages={8589--8601},
  year={2022},
  publisher={IEEE}
}

@inproceedings{lv2021learning,
  title={Learning semantic person image generation by region-adaptive normalization},
  author={Lv, Zhengyao and Li, Xiaoming and Li, Xin and Li, Fu and Lin, Tianwei and He, Dongliang and Zuo, Wangmeng},
  booktitle={Proceedings of the IEEE/CVF conference on computer vision and pattern recognition},
  pages={10806--10815},
  year={2021}
}

@inproceedings{men2020controllable,
  title={Controllable person image synthesis with attribute-decomposed gan},
  author={Men, Yifang and Mao, Yiming and Jiang, Yuning and Ma, Wei-Ying and Lian, Zhouhui},
  booktitle={Proceedings of the IEEE/CVF conference on computer vision and pattern recognition},
  pages={5084--5093},
  year={2020}
}

@inproceedings{pan2024synthesizing,
  title={Synthesizing coherent story with auto-regressive latent diffusion models},
  author={Pan, Xichen and Qin, Pengda and Li, Yuhong and Xue, Hui and Chen, Wenhu},
  booktitle={Proceedings of the IEEE/CVF Winter Conference on Applications of Computer Vision},
  pages={2920--2930},
  year={2024}
}

@article{ramesh2022hierarchical,
  title={Hierarchical text-conditional image generation with clip latents},
  author={Ramesh, Aditya and Dhariwal, Prafulla and Nichol, Alex and Chu, Casey and Chen, Mark},
  journal={arXiv preprint arXiv:2204.06125},
  volume={1},
  number={2},
  pages={3},
  year={2022}
}

@article{xu2021manipulation,
  title={Text guided human image manipulation via image-text shared space},
  author={Xu, Xiaogang and Chen, Ying-Cong and Tao, Xin and Jia, Jiaya},
  journal={IEEE Transactions on Pattern Analysis and Machine Intelligence},
  volume={44},
  number={10},
  pages={6486--6500},
  year={2021},
  publisher={IEEE}
}

@article{xu2021text,
  title={Text-guided human image manipulation via image-text shared space},
  author={Xu, Xiaogang and Chen, Ying-Cong and Tao, Xin and Jia, Jiaya},
  journal={IEEE Transactions on Pattern Analysis and Machine Intelligence},
  volume={44},
  number={10},
  pages={6486--6500},
  year={2021},
  publisher={IEEE}
}

@inproceedings{yu2019vtnfp,
  title={Vtnfp: An image-based virtual try-on network with body and clothing feature preservation},
  author={Yu, Ruiyun and Wang, Xiaoqi and Xie, Xiaohui},
  booktitle={Proceedings of the IEEE/CVF international conference on computer vision},
  pages={10511--10520},
  year={2019}
}

@inproceedings{zhu2017your,
  title={Be your own prada: Fashion synthesis with structural coherence},
  author={Zhu, Shizhan and Urtasun, Raquel and Fidler, Sanja and Lin, Dahua and Change Loy, Chen},
  booktitle={Proceedings of the IEEE international conference on computer vision},
  pages={1680--1688},
  year={2017}
}
\newpage
\appendix

\section{Appendix: Data Processing Details}
\label{sec:appendix_data_processing}

This section details our data standardization and annotation pipeline for Human-Art, OCHuman, and CrowdPose datasets.

\subsection{Pose Standardization}
We unify all skeletal annotations into the COCO-17 format .($17 \times 3$ keypoints)

\textbf{\textit{OCHuman:}} Original COCO-style annotations are converted into instance-level \texttt{.npz} files.

\textbf{\textit{CrowdPose:}} The 14-keypoint annotations are mapped to COCO indices 5--16. Head and neck points without direct COCO equivalents are set to zero to maintain a consistent $17 \times 3$ topology.

\subsection{Multi-granularity Captioning}
We use Qwen3-VL-30B-Instruct to augment training images with multi-scale textual descriptions. 

\textbf{\textit{Inference Settings:}} We employ the vLLM engine with 4-bit AWQ quantization. Images are resized to a maximum side of 1280 pixels while preserving the aspect ratio.

\textbf{\textit{Caption Types:}} For each image, we generate a \textit{short\_caption} (30--60 words) for core attributes and a \textit{long\_caption} (80--180 words) for detailed spatial and pose descriptions.

\textbf{\textit{Style Preservation:}} For artistic datasets like Human-Art, we prepend original style tags (e.g., ``oil painting'') to the VLM-generated prompts.

\subsection{Mixed Training Configuration}
To balance controllability and robustness, we implement a mixed-prompt training strategy.

\textbf{\textit{Mixing Strategy:}} We use a \textit{long\_ratio} of 0.7, meaning 70\% of samples use long captions and 30\% use short captions. Samples are assigned using a fixed random seed.

\textbf{\textit{Dataset Scale:}} The final mixed training set contains \textbf{15,500} images, comprising 8,000 from Human-Art, 5,000 from CrowdPose, and 2,500 from OCHuman.

\textbf{\textit{Format:}} All metadata is stored in a unified JSON mapping, pairing image paths with synchronized captions and pose files.

\section{Implementation details}
\label{subsec:Implementation details}
SD3.5-Medium serves as the pre-trained backbone. In the triple-stream architecture, the pose stream activation layer $L_p$ is set to $12$, and the injection scale $\alpha$ is set to $0.5$. The model is optimized using Adam with a learning rate of $4\times10^{-5}$. A separate learning rate of $10^{-4}$ with FP32 precision optimization is applied to the ten learnable relational bias parameters. For the spatial loss weighting strategy, the threshold $\tau_t$ is set to $0.5$, and $\lambda_{\text{eff}}$ undergoes a linear warmup over the first $500$ steps. Following standard practice, text prompts are randomly dropped with a probability of $0.1$ to strengthen pose guidance. During inference, text prompts are retained, and DDIM sampling with $50$ steps is employed for image generation. The model is trained for 6000 steps on the Human-Art dataset using three NVIDIA A40 GPUs, with bias mask dropout enabled at a probability of $0.2$ to improve generalization.

\section{Evaluation Metrics}
\label{subsec:evaluation_metrics}

To provide a multi-dimensional assessment of TrioPose, we employ six quantitative metrics categorized into pose accuracy, image quality, and semantic alignment. 
\subsection{Pose Accuracy Metrics}
These metrics quantify the structural fidelity of the generated human instances relative to the input pose conditions:

\textbf{\textit{Average Precision (AP):}} Following the standard COCO evaluation protocol, we use a pre-trained ViTPose-H estimator to extract keypoints from the generated images. The AP is calculated by comparing these extracted keypoints against the ground-truth conditional poses using Object Keypoint Similarity (OKS) thresholds.

\textbf{\textit{Cosine Similarity-based Average Precision (CAP):}} CAP measures the angular alignment of limb vectors between the generated and conditional poses. It provides a more nuanced evaluation of joint orientations, which is particularly critical in complex, multi-person overlapping scenarios.

\textbf{\textit{Person Count Error (PCE):}} PCE evaluates the model's ability to generate the exact number of human instances specified by the pose condition. It is defined as the absolute difference between the number of detected persons in the generated image and the number of instances in the input pose map.

\subsection{Image Fidelity and Diversity}
We evaluate the visual realism and distribution alignment of the synthesized results using the following:

\textbf{\textit{Fréchet Inception Distance (FID):}} FID calculates the Fréchet distance between Gaussian distributions fitted to features extracted by a pre-trained Inception-V3 network for both real and generated image sets. Lower FID values indicate higher image fidelity and better coverage of the real data distribution.

\textbf{\textit{Kernel Inception Distance (KID):}} KID measures the squared Maximum Mean Discrepancy (MMD) between Inception representations. Unlike FID, KID is an unbiased estimator, making it more reliable for smaller validation sets. Note that in our quantitative reports, KID values are scaled by $100$ for better readability.

\subsection{Text-Image Semantic Alignment}

\textbf{\textit{CLIP Score:}} This metric utilizes a pre-trained CLIP (ViT-L/14) model to compute the cosine similarity between the latent embeddings of the generated image and the corresponding text prompt. It evaluates the model's success in following complex semantic instructions and text-guided attribute synthesis.

\section{Appendix: Dataset and Model Assets}
\label{sec:appendix_assets}

In accordance with the NeurIPS 2026 guidelines, this section provides detailed documentation for all datasets and pre-trained models utilized in this research. We have strictly adhered to the licenses and terms of use stipulated by the original creators.

\subsection{Detailed Asset Descriptions}
\label{subsec:asset_details}

Our framework and the comparative baselines are trained and evaluated on several distinct assets as described below:

\textbf{\textit{Human-Art}} \cite{ju2023humanart}: The Human-Art dataset comprises 38,000 images distributed across 19 distinct scenarios, encompassing natural scenes, 2D artificial scenarios (e.g., oil paintings, sketches), and 3D artificial scenarios (e.g., statues, digital art). We adopt the standardized train-validation split as suggested by the authors. The dataset is licensed under the Attribution-Non Commercial-Share Alike 4.0 International License (CC BY-NC-SA 4.0).

\textbf{\textit{CrowdPose}} \cite{li2019crowdpose}: This dataset is specifically designed to address human pose estimation in crowded scenes. It contains approximately 20,000 images with 80,000 human instances. The images are categorized by a "crowd index" to reflect varying degrees of occlusion and instance density. It is released under the Creative Commons Attribution 4.0 International License (CC BY 4.0) for academic and non-commercial research purposes.

\textbf{\textit{OCHuman}} \cite{zhang2019ochuman}: The OCHuman dataset focuses on "Occluded Human" instances, providing 5,000 images with over 10,000 instances characterized by extreme overlap and occlusion (at least 50\% of the body). It serves as a rigorous benchmark for multi-instance decoupling. The annotations are licensed under the Creative Commons Attribution-NonCommercial 4.0 International License (CC BY-NC 4.0).

\textbf{\textit{COCO-17}} \cite{lin2014coco17}: As a foundational benchmark, the MS-COCO 2017 dataset provides over 200,000 images across 80 object categories. For our task, we utilize the person category with standard 17-keypoint annotations. The dataset is licensed under the Creative Commons Attribution 4.0 License, posing no specific restrictions for research.

\textbf{\textit{SD3.5-Medium}} \cite{esser2024sd3}: This serves as our pre-trained generative backbone, utilizing a Multimodal Diffusion Transformer (MM-DiT) architecture. The model is released by Stability AI under the Stability Community License, which allows free use for academic research and small-scale commercial applications.

\textbf{\textit{Qwen3-VL}} \cite{bai2025qwen3}: We employ this state-of-the-art vision-language model to augment our training data with multi-granularity textual descriptions. The model and its weights are released under the Apache 2.0 License / Research License, supporting reproducible open-source development.

\textbf{\textit{ViTPose}} \cite{xu2023vitpose}: A Vision Transformer-based pose estimator used in our Pose-Guided Spatial Loss Weighting strategy to extract keypoint heatmaps. The model is available under the Apache 2.0 License.

\subsection{Asset Summary and Licensing}

Table \ref{tab:licenses} provides a quick reference for the licensing information and primary utility of the core assets used in our experiments.

\begin{table}[h]
\centering
\caption{Summary of Dataset and Model Licenses}
\label{tab:licenses}
\small
\begin{tabular}{@{}llll@{}}
\toprule
\textbf{Asset Name} & \textbf{Primary Use} & \textbf{Creator/Source} & \textbf{License Type} \\ \midrule
Human-Art & Benchmark & Ju et al. & CC BY-NC-SA 4.0 \\
CrowdPose & Benchmark & Li et al. & CC BY 4.0 / Academic \\
OCHuman & Benchmark & Zhang et al. & CC BY-NC 4.0 \\
COCO-17 & Pre-training & Microsoft COCO Team & CC BY 4.0 \\
SD3.5-Medium & Backbone Model & Stability AI & Stability Community \\
Qwen3-VL & Data Augmentation & Alibaba Qwen Team & Apache 2.0 \\
ViTPose & Loss Weighting & Xu et al. & Apache 2.0 \\ \bottomrule
\end{tabular}
\end{table}

\subsection{Terms of Use and Compliance}

For all assets listed above, we have ensured compliance with their respective legal frameworks:

\textbf{\textit{Academic and Non-commercial Use:}} Benchmarks such as \textbf{Human-Art} and \textbf{OCHuman} are restricted to non-commercial research purposes. We have verified our usage aligns with these constraints and have not utilized the data for commercial product development.

\textbf{\textit{Model Redistribution:}} The \textbf{SD3.5-Medium} model is governed by the Stability Community License, which permits free use for research. Our academic publication and proposed release of derived weights fall within these permitted boundaries.

\textbf{\textit{Standardized Protocol:}} All datasets used in this research adhere to a standardized keypoint format featuring 17 keypoints per person, consistent with the COCO and Human-Art protocols. 

\textbf{\textit{Data Integrity:}} We have not re-distributed raw dataset files; instead, we provide scripts to acquire data directly from original sources to respect creators' distribution rights.


\end{document}